\documentclass[runningheads]{llncs}
\usepackage{graphicx}
\usepackage{cite}
\usepackage{hyperref}
\usepackage{amsmath}
\usepackage{booktabs}

\begin{document}
    \title{Using Features at Multiple Temporal and Spatial Resolutions to Predict Human Behavior in Real Time\thanks{This research was conducted as part of DARPA's Artificial Social Intelligence
    for Successful Teams (ASIST) program, and was sponsored by the Army Research
    Office and was accomplished under Grant Number W911NF-20-1-0002. The views and
    conclusions contained in this document are those of the authors and should not
    be interpreted as representing the official policies, either expressed or
    implied, of the Army Research Office or the U.S. Government. The U.S.
    Government is authorized to reproduce and distribute reprints for Government
    purposes notwithstanding any copyright notation herein.}}
%
%
    \author{Liang Zhang\inst{1}\orcidID{0000-0002-1966-0864} \and
    Justin Lieffers\inst{1}\orcidID{0000-0003-4527-8850} \and
    Adarsh Pyarelal\inst{1}\orcidID{0000-0002-1602-0386}}
    \authorrunning{L. Zhang et al.}
%
    \institute{University of Arizona, Tucson AZ 85721, USA
    \email{\{liangzh,lieffers,adarsh\}@email.arizona.edu}}
    \maketitle              
    \begin{abstract}
        When performing complex tasks, humans naturally reason at multiple temporal and spatial resolutions simultaneously.
        We contend that for an artificially intelligent agent to effectively model human teammates, i.e., demonstrate computational theory of mind (ToM), it should do the same.
        In this paper, we present an approach for integrating high and low-resolution spatial and temporal information to predict human behavior in real time and evaluate it on data collected from human subjects performing simulated urban search and rescue (USAR) missions in a Minecraft-based environment.
        Our model composes neural networks for high and low-resolution feature extraction with a neural network for behavior prediction, with all three networks trained simultaneously.
        The high-resolution extractor encodes dynamically changing goals robustly by taking as input the Manhattan distance difference between the humans' Minecraft avatars and candidate goals in the environment for the latest few actions, computed from a high-resolution gridworld representation.
        In contrast, the low-resolution extractor encodes participants' historical behavior using a historical state matrix computed from a low-resolution graph representation.
        Through supervised learning, our model acquires a robust prior for human behavior prediction, and can effectively deal with long-term observations.
        Our experimental results demonstrate that our method significantly improves prediction accuracy compared to approaches that only use high-resolution information.

    \end{abstract}

    \section{Introduction}
    Artificially intelligent (AI) teammates should have a number of capabilities to
    be effective \cite{seeber2020machines}, including inferring the
    internal states of other agents \cite{Rabinowitz:2018, Baker:2017, Wu:2018},
    solving problems collaboratively with them \cite{Galescu:2018, Allen:2018,
        Perera:2018}, and communicating with them in a socially-aware manner
    \cite{DBLP:conf/iva/ZhaoPC14, DBLP:conf/iva/ZhaoSBC16}. While these
    capabilities have been developed to some extent for simple domains (e.g., 2D
    gridworlds) and simulated agents, current state of the art approaches still
    face significant challenges when it comes to dealing with complex domains and
    modeling actual human teammates (as opposed to simulated agents). We attempt
    to address some of these challenges in the context of an experiment involving
    humans conducting a simulated urban search and rescue (USAR) mission set in a
    Minecraft-based environment \cite{Huang.ea:2021}.


    This domain is significantly more complex than the domains previously studied
    in the literature on computational theory of mind (ToM) \cite{Baker:2017,
        Rabinowitz:2018}. Enabling AI agents to understand human behavior in complex
    domains will be essential to achieve the goal of better human-AI teaming. The
    complexity of the domain and the emphasis on analyzing human subjects lead to a
    few unique challenges, which we describe below.

    \begin{itemize}

        \item \textbf{Limited data.} Since collecting human subjects data is
        expensive and time-consuming, the amount of training data available to
        us is very limited. This rules out using certain classes of modern
        machine learning approaches (e.g., transformer architectures) that
        require a large amount of training data.

        \item \textbf{Noisy data.} Human subjects data is typically noisy,
        especially in the short term, with participants frequently violating
        assumptions of rationality that are used in existing works on
        computational ToM \cite{Baker:2017, Rabinowitz:2018, DBLP:conf/nips/Zhi-XuanMSTM20}. This expresses the need for a two resolution
        approach as rationality can often be recovered when the domain is represented
        at a lower resolution and the noise is averaged over, yet the high resolution is required
        for real-time predictions.

        \item \textbf{Long horizon.} In contrast to earlier works on computational
        ToM \cite{Baker:2017, Rabinowitz:2018, DBLP:conf/nips/Zhi-XuanMSTM20}
        that study domains with $\approx10^2$ primitive actions per
        episode\footnote{We use the term \emph{episode} to denote a sequence of
        actions taken by an agent to perform a given task. We also use the
        term \emph{trial} elsewhere in the paper to denote the same thing.},
        our work considers a domain with episodes containing $\approx10^3$
        primitive actions and a far larger observation space including more
        than 20 areas and complex connectivity. This requires us to implement
        a long-term memory mechanism and the ability to extract key features
        from large amounts of noisy data, both of which are challenging in
        their own right.

        \item \textbf{Complex dynamics.} Our domain is large and possesses a
        complex topological structure, coupled with a complex rescue mechanism
        setting (for details, please refer to the approach section), which
        require us to consider human behavior at different levels of spatial
        and temporal granularity. Our model simultaneously takes into account
        both the \emph{short-term goal preference} in a local area and the
        \emph{long-term rescue strategy}.

    \end{itemize}

    To address the above challenges, we propose a two-level representation. The
    first is a \emph{low-resolution} level that contains information about the
    topology of the environment (i.e. which areas are connected to each other) and
    the status of victims in each area. The second is a \emph{high-resolution}
    level that contains more granular information about the environment, such as
    the Cartesian coordinates corresponding to the agent's current location, walls,
    openings, and victims inside the rooms.

    For the low-resolution representation, we build a matrix that encodes key
    historical information, which helps our model learn high-level features of
    human behavior, such as long-term search and rescue strategies. In contrast,
    for the high-resolution representation, we organize the input vector to our
    proposed model based on the latest short-term observations, which are more
    conducive to recognizing short-term goal preferences. Using the high and low
    level resolutions simultaneously aligns with the way humans reason about
    complex tasks, and also results in better performance on our prediction tasks,
    compared to considering only a single resolution.

    \section{Related Work}

    There exist a number of other approaches to computational ToM in the
    literature. In this section, we describe some of them, along with their
    advantages and disadvantages compared with our approach.

    Bayesian Theory of Mind (BToM) models \cite{DBLP:conf/cogsci/BakerST11,
        Baker:2017, Jara-Ettinger.ea:2020} calculate the probabilities of potential
    goals of an agent and other's beliefs. These models are primarily
    based on Markov Decision Process (MDP) formalisms and thus suffer from high computational costs
    for complex domains.

    Zhi-Xuan et al. \cite{DBLP:conf/nips/Zhi-XuanMSTM20} proposed an online Bayesian goal
    inference algorithm based on sequential inverse plan search (SIPS). This approach
    allows for real-time predictions on a number of different domains. Notably,
    their approach models agents as \emph{boundedly rational planners}, thus making them
    capable of executing sub-optimal plans, similar to humans. However, this
    approach cannot be directly applied to our domain due to the fact that our
    agents (i.e., humans) have incomplete knowledge of their domain and thus the
    short term planning would suffer without added hierarchical complexity
    or longer term planning. In our proposed approach, we use a similar idea of
    calculating the probabilities of potential goals, but we use neural networks
    which allow for the automatic extraction of features and correlations from the
    data without having to hand-craft conditional probability distributions.

    Our supervised learning approach considers both long-term historical and
    real-time high-resolution features in a robust fashion, dramatically reducing
    the computational costs of training and deployment in online settings even for
    complex domains.

    Inverse reinforcement learning (IRL) methods \cite{DBLP:conf/icml/NgR00,
        DBLP:conf/icml/PieterN04, DBLP:conf/ijcai/RamachandranA07,
        DBLP:journals/corr/Hadfield-Menell16} make real-time predictions about an agent
    from learning the agent's reward function by observing its behavior. However,
    IRL methods suffer in online settings for complex domains because they are
    based on MDP formalisms, similar to BToM approaches
    \cite{DBLP:journals/corr/abs-1912-04472, DBLP:conf/icra/MichiniH12}.

    Approaches based on plan recognition as planning (PRP), which use classical
    planners to predict plan likelihoods given potential goals, can also give
    real-time predictions for complex domains \cite{DBLP:conf/ijcai/RamirezG09,
        DBLP:conf/aaai/RamirezG10, DBLP:conf/ijcai/SohrabiRU16,
        DBLP:conf/aaai/HollerBBB18, DBLP:conf/aaai/KaminkaVA18,
        DBLP:conf/aaai/VeredPMMK18}. However, these methods require labor-intensive
    manual knowledge engineering, which can be prohibitive for environments that
    have complex dynamics. Additionally, these methods struggle with the noisy and
    sub-optimal nature of human behavior. In contrast, our neural network based
    approach requires minimal manual knowledge engineering and our two levels of
    resolutions allow for an effective treatment of noisy/sub-optimal behavior.

    Guo et al. \cite{Guo:2021} study the same domain as the one in this paper, and use a
    graph-based representation for their model as well. However, they focus on
    transfer learning as a way to improve training when dealing with a limited
    amount of training data. Additionally, their agent predictions are focused on
    naviagtion. The techniques developed in their work are applicable to us and
    could be useful to further expand our model in the near future.

    Lastly, Rabinowitz et al. \cite{Rabinowitz:2018} used meta-learning to build models of the
    agents from observations of their behavior alone. This resulted in a
    prior model for the agents' behavior and allowed for real-time
    predictions. However, this approach only studied situations where the agents
    followed simple policies, and the dynamics of their domain are much simpler
    than ours.

    \section{Approach}

    \subsection{Domain and task}

    \begin{figure}
        \centering
        \includegraphics[width=0.8\textwidth]{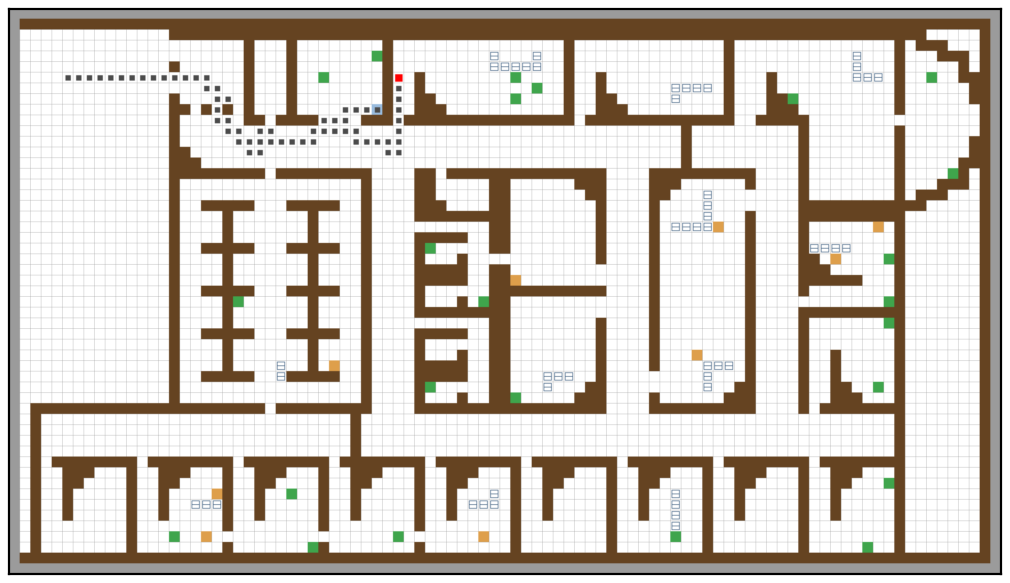}
        \caption{
            A visualization of the high-resolution representation for our domain.
            The red dot represents the agent (i.e., the human's Minecraft avatar),
            and the grey dots represent grid cells that the agent has traversed in
            the past. Green and yellow squares represent untriaged victims, blue
            squares represent triaged victims, brown squares represent walls, and
            grey squares represent obstacles. Walls and obstacles are not
            traversable, and the blank (white) squares are walkable areas.
        }
        \label{universe}
    \end{figure}

    The domain we consider is that of a USAR mission simulated in a Minecraft-based
    environment \cite{Huang.ea:2021}. In this scenario, the participants must
    navigate an office building that has suffered structural damage and collapse
    due to a disaster. The original building layout is altered by the collapse,
    with some passages being closed off due to rubble, and new openings being
    created by walls collapsing.

    The goal of the mission is to obtain as many points as possible by triaging
    victims of the building collapse within a 10-minute time limit. There are 34
    victims in the building, among whom 10 are seriously injured and will expire 5
    minutes into the mission. These \emph{critically injured} victims take 15
    seconds to triage and are worth 30 points each. These victims are represented
    by \emph{yellow} blocks. The other victims are considered \emph{non-critically
    injured}, take 7.5 seconds to triage, and are worth 10 points each. These
    victims are represented by \emph{green} blocks.

    Each participant conducts three versions of the mission, with different levels
    of difficulty (easy, medium, and hard). On higher difficulty levels, the
    victims are less clustered, further away from the starting point, and are more
    difficult to find. Higher difficulty levels also have more alterations from the
    original static map that the participants are provided at the beginning of the
    mission (i.e., more blockages and openings).

    \subsection{Representation}

    \subsubsection{High-Resolution Representation}

    We use a highly simplified 2D gridworld environment representation for the high
    resolution representation. In this representation, we encode different objects
    and store them in a $51\times91$ integer matrix. The specific encodings are
    shown in Table~\ref{object_encoding}.

    \begin{table}[h]
        \centering
        \caption{Encodings for objects in the high resolution representation.}
        \label{object_encoding}
        \begin{tabular}{|l|l|}
            \hline
            \toprule
            Object                                  & Value \\\midrule
            \hline
            Empty                                   & 1     \\
            Wall                                    & 4     \\
            Critical victim                         & 81    \\
            Non-critical victim                     & 82    \\
            Unavailable victim (triaged or expired) & 83    \\
            Obstacle                                & 255   \\
            Agent                                   & 0     \\
            \bottomrule
            \hline
        \end{tabular}

    \end{table}

    In Fig.~\ref{universe}, we show a visualization of the high-resolution
    representation\footnote{Our high-resolution visualization code implementation
    is based on this repository: \url{https://gitlab.com/cmu_asist/gym_minigrid}}.
    Our primitive action space consists of two types of actions: \emph{move} and
    \emph{triage}. The `move' action can be carried out in four directions: up,
    down, left, and right, moving one cell at a time when the direction of moving
    is not obstructed. The `triage' action can only be performed when the agent
    reaches locations cells where victims are located.

    In this high-resolution representation, we can analyze human behavior based on
    discrete primitive actions combined with the layout of the building, which
    enables modeling real-time changes in short-term goal preferences. However,
    these actions also introduce noise, and inference based on them alone is not
    conducive to extracting high-level features and organizing long-term memory due
    to the large number of primitive actions per trial ($\approx 10^3$).

    \subsubsection{Low-Resolution Representation}

    To facilitate the extraction of high-level features from human behavior and the
    organization of long-term historical information, we construct a graph-based
    representation to simplify our domain further. The nodes of the represent areas
    (e.g., rooms, hallways, etc.) of the building, the edges represent connections
    between areas, and each node has three integer-valued attributes:

    \begin{itemize}
        \item Number of green victims in the area.
        \item Number of yellow victims in the area.
        \item Visited status. This attribute can take one of four possible values:
        \begin{itemize}
            \item 0: The node has not been previously visited by the agent.
            \item 1: The node has been previously visited by the agent.
            \item 2: The agent is currently located at the node.
            \item 3: The node was the previous node the agent was at.
        \end{itemize}
    \end{itemize}

    \begin{figure}
        \centering
        \includegraphics[width=0.6\textwidth]{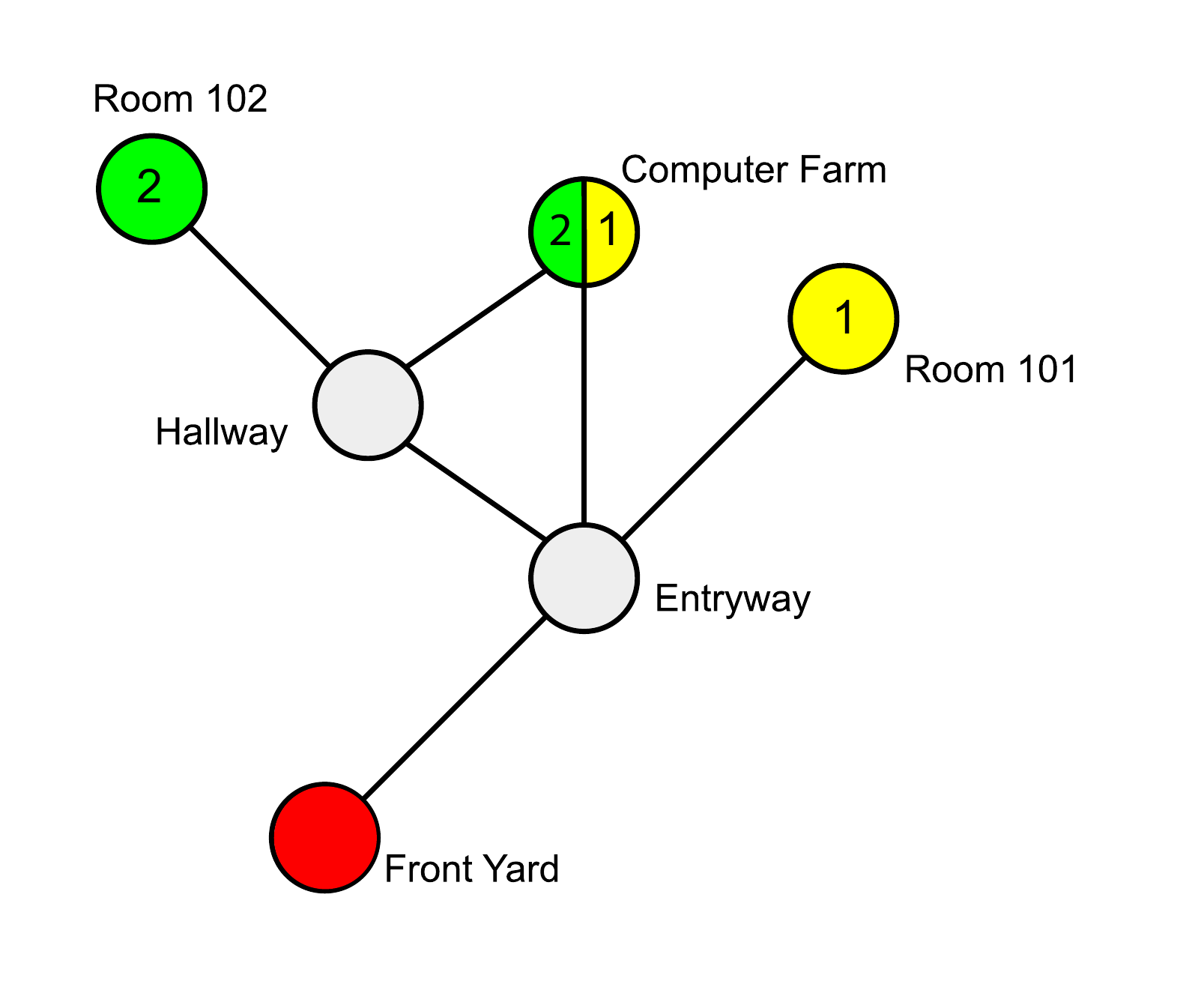}
        \begin{tabular}{|l|c|c|c|}
            \toprule
            \hline
            Area ID       & Yellow victims & Green victims & Visited status \\
            \hline
            \midrule
            Room 101      & 1              & 0             & 0              \\
            Room 102      & 0              & 2             & 1              \\
            Front Yard    & 0              & 0             & 2              \\
            Entryway      & 0              & 0             & 3              \\
            Hallway       & 0              & 0             & 1              \\
            Computer Farm & 1              & 2             & 0              \\
            \bottomrule
            \hline
        \end{tabular}
        \caption{
            Visualization of an example low resolution graph representation and the
            corresponding memory matrix. The nodes represent the areas in the
            building, and the edges the connections between them. The number and
            type of victims in each area are recorded as attributes on each node,
            and are shown using a color and number indicating the type and quantity
            of victims. The red node represents the node the agent is in. The
            matrix below the graph is the corresponding low resolution memory
            matrix.
        }
        \label{sample}
    \end{figure}

    For `visited status' attribute, if two conditions are met at the same time, the
    higher encoding value has a higher priority. For example, if the agent returns
    to a previously visited room, the visited status of the current room defaults
    to 2 instead of 1 even though both are applicable. The visited status in the
    memory matrix is updated according to the above rules when the agent moves from
    one area to the next.
    In addition, when the agent successfully triages a specific type of victim, the number of victims of that type in the current area is reduced by one.
    Therefore, the updates to this matrix record the historical behavior of the current agent.

\begin{figure}
        \centering
        \includegraphics[width=0.8\textwidth]{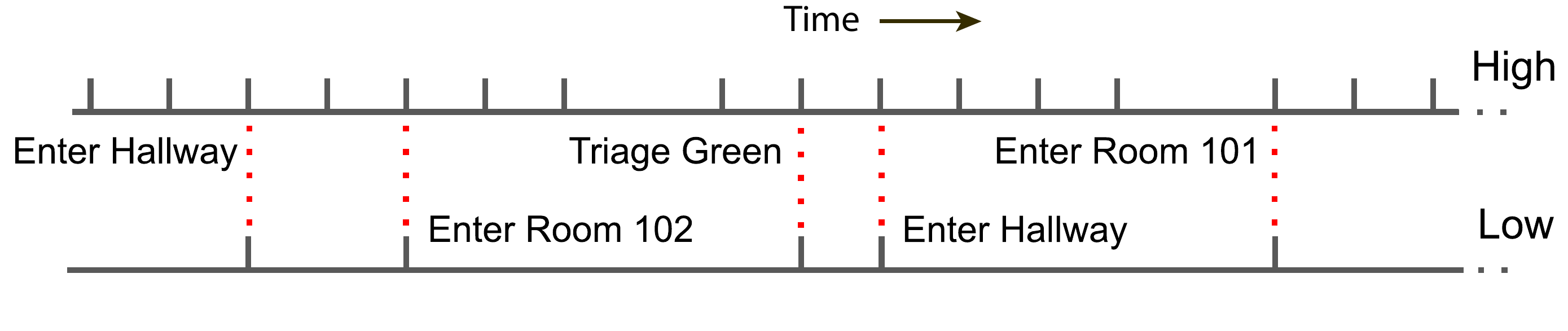}
        \caption{
            An example sketch showing the different time intervals of state
            updating for the two resolutions. Each tick line indicates an update
            to the state, and the red dotted lines connect ticks with the same
            timestamp The high resolution input is updated for every primitive
            action, while the low resolution input is only updated when the agent
            leaves a node or changes the attributes of a node (triaging a victim),
            hence the lesser number of ticks.
        }
        \label{timeline}
    \end{figure}

    This is a dramatic simplification of our domain, since we ignores many details from the environment, such as the specific locations of agents and victims, the detailed layout of the building, etc.
    Therefore, the low-resolution representation provides a more concise encoding of crucial
    historical information, making it easier for the model to extract high-level
    features in human behavior. We organize this information into a matrix.
    However, the time interval for state updates is longer than that in the
    high-resolution representation, since we are not encoding primitive actions for
    this representation, and it cannot grasp real-time changes in human intentions.
    Figure~\ref{timeline} shows an example sketch of the time intervals for
    updates to the state in the two resolutions.
    In order to leverage the complementary strengths of these two resolutions, we
    propose a model that uses both as inputs simultaneously.

    \subsection{Model}

    Our model produces two types of outputs: (i) \emph{goals}, i.e.,
    objects/locations that the agent is trying to get to, and (ii) the \emph{next
    type of victim} (green or yellow) that the agent will attempt to triage.

    \subsubsection{Goals}

    \begin{figure}
        \centering
        \includegraphics[width=0.6\textwidth]{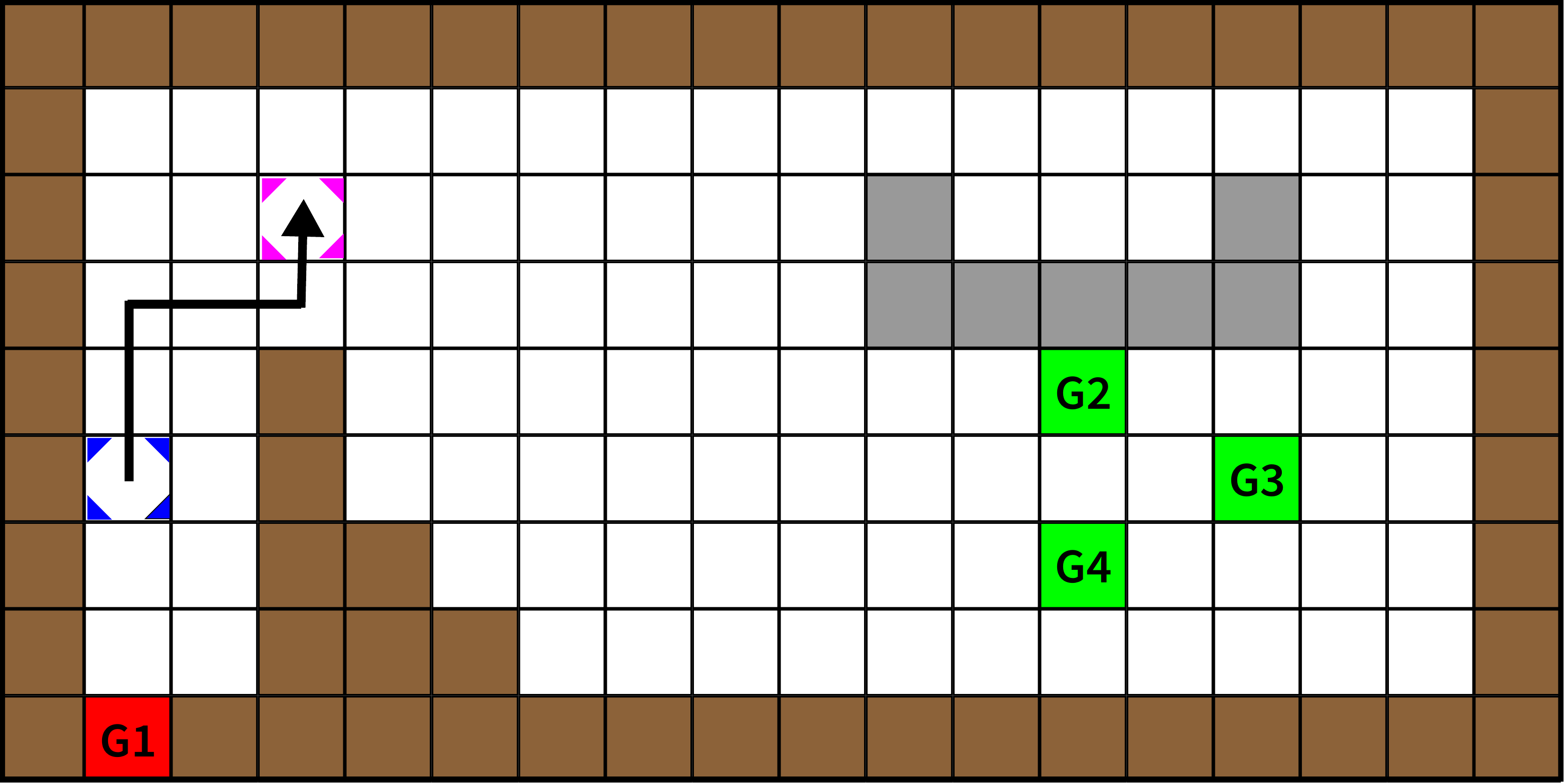}

        \begin{tabular}{|l|r|r|}
            \hline
            \toprule
            Goal & $\Delta_\text{MD}$ & Likelihood \\
            \hline
            \midrule
            G1   & -5                 & 0.4298     \\
            G2   & 3                  & 0.2075     \\
            G3   & 3                  & 0.1813     \\
            G4   & 3                  & 0.1814     \\
            \bottomrule
            \hline
        \end{tabular}

        \begin{tabular}{c}
            \hline
            \toprule
            \hline
            Probability that the next victim to be triaged is yellow \\ \midrule
            \hline
            0.8994                                                   \\
            \bottomrule
            \hline
        \end{tabular}

        \caption{
            An example of how we deal with the data from the high resolution
            representation. The arrows represent the agent's last six movements.
            The quantity $\Delta_\text{MD}(g, 6)$ (see eq.~\ref{mdd}) is computed
            for $g\in\left\{G1, G2, G3, G4\right\}$ and shown in the table
            below the figure, along with the predictions of the model for each
            potential goal $g$ in the area. The window of `move' actions is from
            when the agent moves from the magenta outlined square to the blue
            outlined square.
        }
        \label{example}
    \end{figure}

    The primary outputs of our model are similar to methods based on Bayesian ToM
    approaches \cite{DBLP:conf/cogsci/BakerST11, Baker:2017,
        Jara-Ettinger.ea:2020}. We consider victims and portals connecting adjacent
    areas as potential goals, and aim to predict which goal the agent is currently
    pursuing. See Figure~\ref{example} for an example set of goals available to an
    agent when entering a particular room.

    \subsubsection{Next Triaged Victim Type}

    In addition to predicting the probability of the agent pursuing a potential
    goal, we also predict the type of victim to be triaged next, which helps us
    identify the agent's strategy or long term behavior. For example, we observed
    that some players prioritize triaging yellow victims because they are worth
    more points and expire sooner, while some players are more opportunistic,
    triaging victims in the order they appear in their field of view. Note that the
    next victim to be triaged may not be in the current area that the player is in.
    Thus, we need to leverage information from both the high and low resolution
    representations to make this prediction, making it an important output that
    takes advantage of our multi-resolution architecture.

    \subsection{Architecture}

    \begin{figure}
        \includegraphics[width=\textwidth]{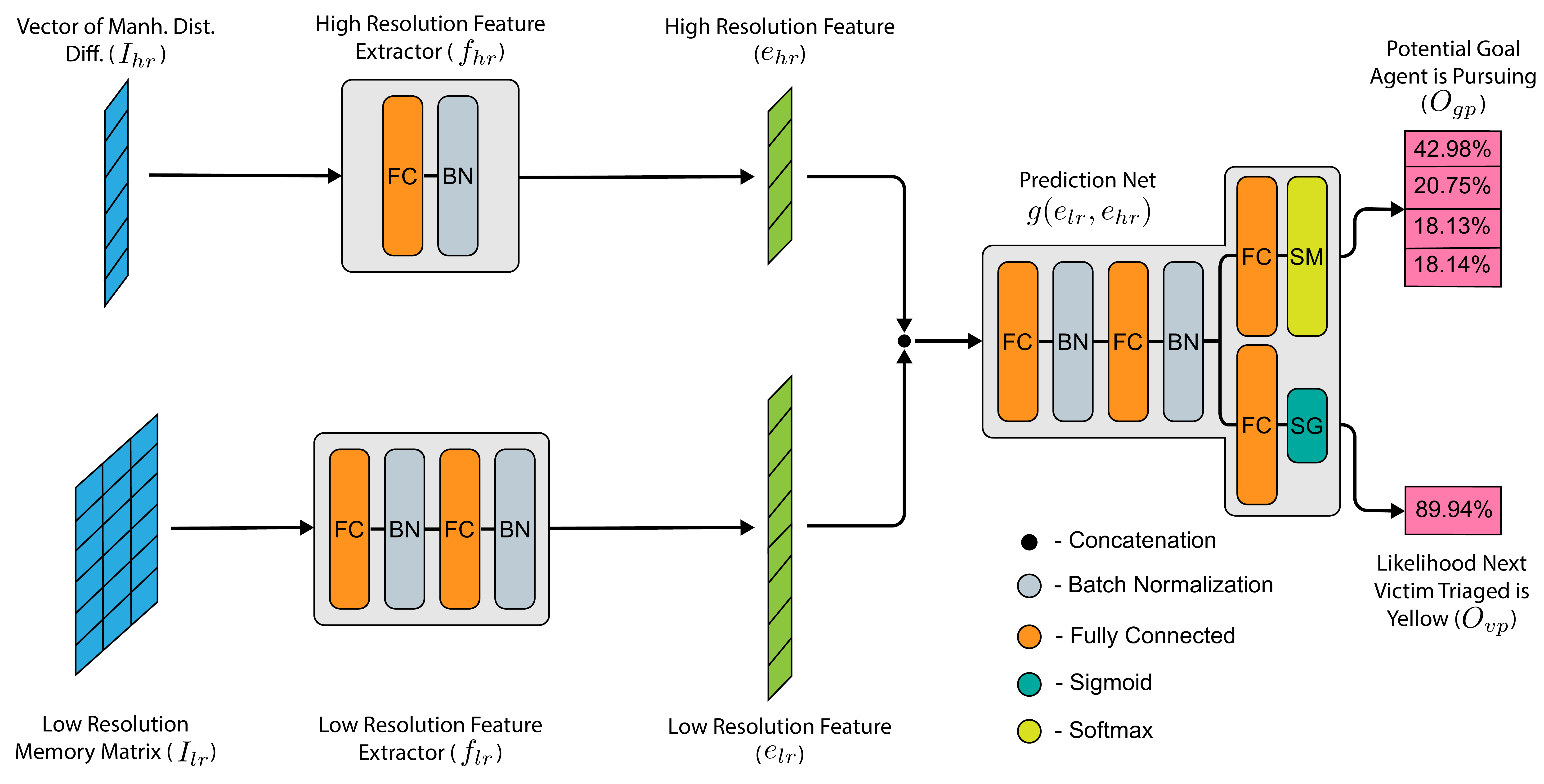}
        \caption{
            This is our network architecture. Our inputs are fed into features
            extractors for each resolution and then those extracted features are
            concatenated and fed into the prediction net which produces our goal
            and victim type predictions. The values shown for $O_{gp}$ and $O_{vp}$
            are taken from the example in Fig.~\ref{example} for illustrative
            purposes.
        }
        \label{network}
    \end{figure}

    The architecture of the model is shown in Figure~\ref{network}. First, the
    information from the high and low resolution representations are used as
    inputs. The high-resolution input \(I_{hr}\) is a vector of $\Delta_\text{MD}$
    values, one for each goal. The low-resolution input \(I_{lr}\) is the memory
    matrix described earlier. The corresponding features \(e_{hr} =
    f_{hr}(I_{hr})\) and \(e_{lr} = f_{lr}(I_{lr})\) are extracted by the feature
    extractor networks \(f_{hr}\) and \(f_{lr}\), respectively. Then, these two
    features from the two different resolutions are concatenated and fed into the
    prediction net $g$. The next goal and victim type to be triaged predictions
    \(O_{gp}\) and \(O_{vp}\) take the form of estimating the two probabilistic
    outputs with \(g(e_{hr}, e_{lr})\).
    Since the inputs consider state differences rather than the entire state, the
    size of the input observation space is significantly reduced, thereby reducing
    the training difficulty of our deep learning model.
    We use a fully connected (FC) layer combined with a batch
    normalization layer as a basic building block for our three neural networks.
    The output FC layers in the prediction network ($g(e_{lr}, e_{hr})$) are passed through
    softmax and sigmoid functions to obtain the probabilities of the agent's goal
    ($O_{gp}$) and the likelihood that the next victim is triaged ($O_{vp}$), respectively.

    \subsubsection{High-Resolution Input}

    Similar to the setting of the BToM \cite{Baker:2017}, we infer the probability of pursuing a
    goal. As shown in Figure~\ref{example}, we compute the quantity
    $\Delta_\text{MD}$, defined as follows:

    \begin{equation}
        \Delta_\text{MD}(g, m) =
        D(x_{i}^m, x_g) - D(x_{f}^m, x_g)
        \label{mdd}
    \end{equation}

    \noindent where $x_{i}^m$ and $x_{f}^m$ are the initial and final positions of
    the agent computed with respect to a window of the past $m$ `move' primitive
    actions, $x_g$ is the location of the goal $g$ for which $\Delta_\text{MD}$ is
    being calculated, and $D(a,b)$ is the Manhattan distance between locations $a$
    and $b$. We found that setting $m = 6$ to be the best fit choice, which still
    gives real-time predictions, while also handling some noise in the agent's
    actions. See Table~\ref{results2} for a comparison of results with different
    values of $m$.

    \subsubsection{Low-Resolution Input}

    As shown in Figure~\ref{sample}, we record the victim status and area
    visitation status of each area in a matrix and use it as an input to the
    proposed model. This input helps us extract long-term historical information
    to form memory and facilitate the extraction of high-level features (long term
    strategies) as a prior to human behavior predictions.

    \section{Evaluation}

    Our model is trained in an end-to-end manner via supervised learning using an
    Nvidia V100S GPU and the Adam optimization algorithm \cite{Kingma.ea:2015}. We
    calculate the softmax cross entropy loss for goal prediction and the binary
    cross entropy loss for victim type prediction. The training loss
    \(L_\text{total}\) is the sum of the goal prediction loss \(L_\text{gp}\) and
    the victim type loss \(L_\text{vp}\) as seen in eq.~\ref{tl}, where the victim
    type loss weight, \emph{W}, is given in Table~\ref{hyper_para}, along with the
    rest of the training hyperparameters after tuning.

    \begin{equation}
        L_\text{total} = L_\text{gp} + W * L_\text{vp}
        \label{tl}
    \end{equation}

    Figure~\ref{example} illustrates how our proposed model works. As shown in
    Fig.~\ref{universe}, in the room that the agent searched just prior to the room
    that it is currently in, the agent only triaged the yellow victim and left the
    two green victims, which hints that that the agent is likely following a
    strategy that prioritizes rescuing all the yellow victims first. Our model
    encodes this behavior as prior knowledge and predicts that the probability that
    the next victim to be triaged will be yellow is $\approx0.9$.

    \begin{table}[h]
        \centering
        \caption{Hyperparameters for our model training.}
        \label{hyper_para}
        \begin{tabular}{|l|r|}
            \hline
            \toprule
            Hyperparameter                     & Value \\\midrule
            \hline
            Learning rate                      & 0.001 \\
            Low resolution feature size        & 64    \\
            High resolution feature size       & 4     \\
            Hidden size for prediction net     & 64    \\
            Batch size                         & 16    \\
            Random seed                        & 0     \\
            Victim type loss weight (\emph{W}) & 0.3   \\
            \bottomrule
            \hline
        \end{tabular}

    \end{table}

\begin{figure}
        \centering
        \includegraphics[width=0.6\textwidth]{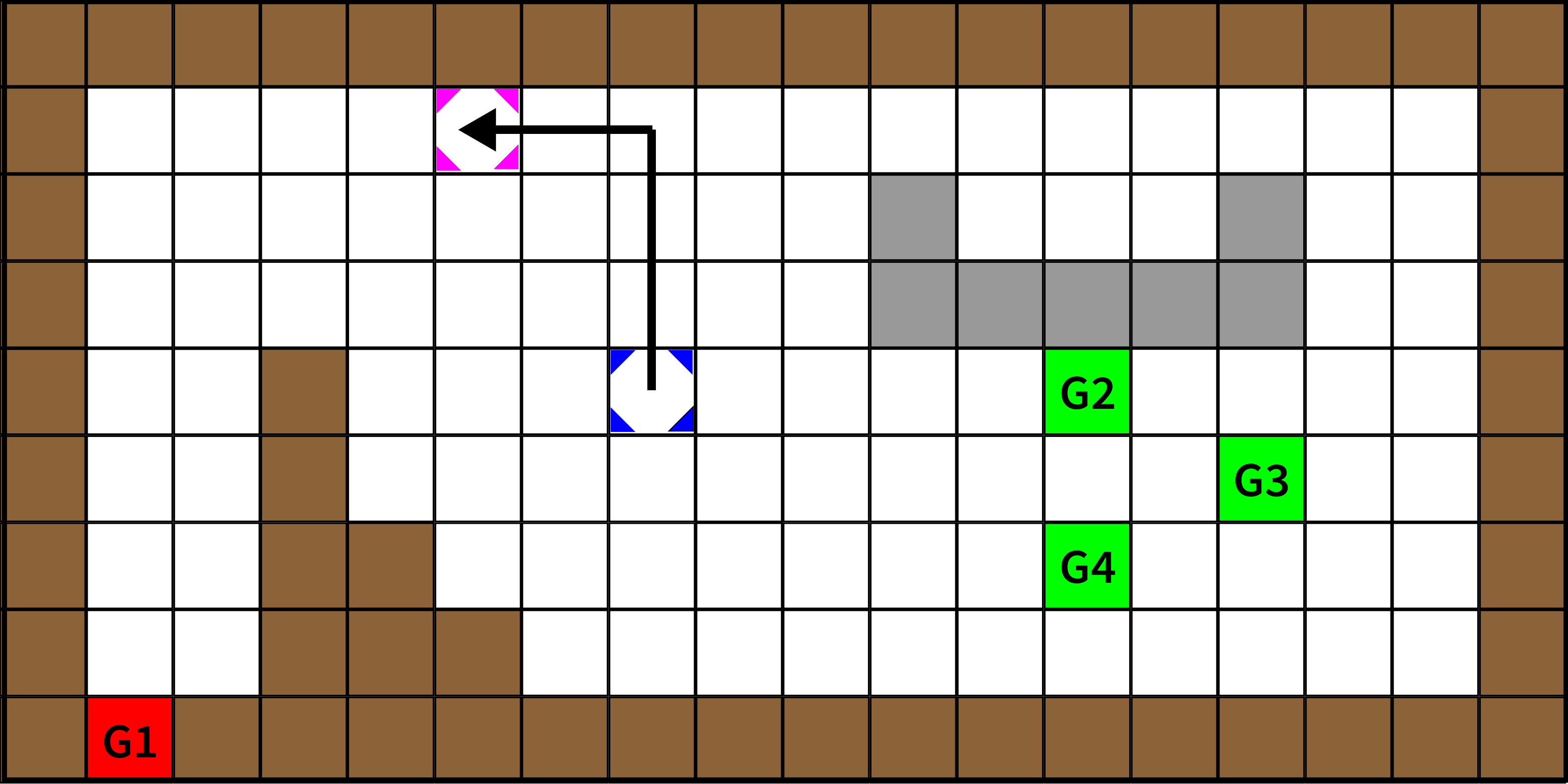}

        \begin{tabular}{|l|r|r|}
            \hline
            \toprule
            Goal & $\Delta_\text{MD}$ & Likelihood \\
            \hline
            \midrule
            G1   & 1                  & 0.4674     \\
            G2   & -5                 & 0.1804     \\
            G3   & -5                 & 0.1761     \\
            G4   & -5                 & 0.1761     \\
            \bottomrule
            \hline
        \end{tabular}

        \begin{tabular}{c}
            \hline
            \toprule
            \hline
            Probability that the next victim to be triaged is yellow \\ \midrule
            \hline
            0.9447                                                   \\
            \bottomrule
            \hline
        \end{tabular}

        \caption{
            The high resolution representation at a later time than the example
            shown in Fig.~\ref{example}. Here we see the probabilities of the agent
            heading to goal G1, and that the next victim triaged will be yellow are
            increasing, showing that our model is correctly predicting the agent's
            goals in real-time, in addition to showing our prediction at an earlier
            timestep was correct.
        }
        \label{example_2}
    \end{figure}

    Without a prior about the rescue strategy we may naively expect that the agent
    will move from G2 to G3 or G4 (i.e., to the next closest victim) with a high
    probability. In contrast, our model predicts that the most likely next
    short-term goal is the room's exit, with a probability of $\approx 0.43$. In
    Figure~\ref{example_2} (similar to Fig.~\ref{example}), the same player finally
    chose to leave after finding there is no yellow victims in this room. The
    probability of the agent returning to G1 to try and find a yellow victim to
    triage next can be seen to increase from about 0.90 to 0.95. This
    demonstrates that our model can learn about high-level strategies that a player
    is following, and can also detect instantaneous changes in the short-term goals
    of human players.

    \begin{table*}
        \centering
        \caption{
            Results for 6-fold cross-validation for our approach and two baselines
            based on high-resolution inputs. In the first method, we encode the
            high resolution input as the destination locations of the agent's most
            recent six `move' actions. For the second method, we concatenate the
            high resolution input vector $I_{hr}$ with an integer representing
            which area the agent is in. We find that our multi-resolution approach
            significantly outperforms the baselines that only use high resolution
            inputs.
        }
        \begin{tabular}{|l|r|r|r|r|r|r|}
            \hline
            \toprule
            & \multicolumn{2}{|c|}{Easy} & \multicolumn{2}{|c|}{Medium} & \multicolumn{2}{|c|}{Hard} \\
            \hline
            Model - Cross Val.          & Goal Acc.       & Vic. Acc.       & Goal Acc.       & Vic. Acc.       & Goal Acc.       & Vic. Acc.       \\\midrule
            \hline
            High Res. (Locations)       & 0.6313          & 0.7060          & 0.6232          & 0.6874          & 0.6031          & 0.6838          \\
            High Res. ($\Delta_{MD}^i$) & 0.6526          & 0.7315          & 0.6412          & 0.7037          & 0.6251          & 0.6881          \\
            High + Low Res.             & \textbf{0.7208} & \textbf{0.9008} & \textbf{0.7146} & \textbf{0.8803} & \textbf{0.6780}    & \textbf{0.8881}          \\
            \bottomrule
            \hline
        \end{tabular}

        \label{results}
    \end{table*}

    In Table~\ref{results}, we compare our multi-resolution method to two baseline
    approaches based solely on high-resolution information\footnote{We do not
    compare with an approach based solely on low-resolution information, as it
    would be not be sufficient to differentiate between multiple short-term
    goals within a single area/node}. The first baseline uses the 2D coordinates of
    the destination cells of the six most recent `move' actions as the input. The
    second baseline considers the high-resolution input based on $\Delta_\text{MD}$
    and includes only a small portion of the information from the low-resolution
    representation. Specifically, since the current area cannot be encoded if only
    $\Delta_\text{MD}$ is considered, we encode each area with a unique integer and
    append this integer to the input vector $I_{hr}$.

    We have 66 trials for each difficulty level, and use a 6-fold cross-validation
    procedure to evaluate our model\footnote{We evaluate the accuracy of the victim
    type prediction only in the first five minutes of each trial because yellow
    victims expire after five minutes, leaving only green victims to triage}.
    As shown in Table~\ref{results}, we see that the baseline using
    $\Delta_\text{MD}$ performs better than the baseline that uses only the
    past six destination cells of the agent's `move' actions, and our approach
    that uses both high and low resolution information outperforms both the
    baselines. Compared to using location information alone, using
    $\Delta_\text{MD}$ (or more specifically, a vector of $\Delta_\text{MD}$
    values, one for each goal, i.e., $I_{hr}$) as an input can lead to better
    features being extracted, thus improving prediction accuracy. Our proposed
    method based on the combination of high and low-resolution information
    allows our model to effectively learn the relationship between features at
    multiple resolutions in the data, further improving the accuracy of
    behavior prediction.

    \begin{table*}
        \centering
        \caption{
            Results for 6-fold cross-validation for our approach in which
            the high resolution inputs are based on different numbers of `move' actions.
        }
        \label{results2}
        \begin{tabular}{|c|r|r|r|r|r|r|}
            \hline
            \toprule
            & \multicolumn{2}{|c|}{Easy} & \multicolumn{2}{|c|}{Medium} & \multicolumn{2}{|c|}{Hard} \\
            \hline
            last \textit{m} moves & Goal Acc.       & Vic. Acc.       & Goal Acc.       & Vic. Acc.       & Goal Acc.       & Vic. Acc.       \\\midrule
            \hline
            3                     & 0.7181          & \textbf{0.9037} & 0.7071          & 0.8816          & 0.6712          & 0.8857          \\
            6                     & \textbf{0.7208} & 0.9008          & \textbf{0.7146} & 0.8803          & 0.6780          & \textbf{0.8881} \\
            12                    & 0.7151          & 0.9001          & 0.7118          & \textbf{0.8835} & \textbf{0.6801} & 0.8845          \\
            \bottomrule
            \hline
        \end{tabular}

    \end{table*}

    We also investigated the sensitivity of our approach to the choice of the
    parameter $m$ (the number of moves in our window when we compute
    $\Delta_\text{MD}(m, g)$).
    The results are shown in Table~\ref{results2}.
    The performance of our proposed method is not overly sensitive to the number of
    moves, and thus we choose $m=6$ after comprehensively considering the results for the three tasks.

    \section{Conclusion}
    In this paper, we proposed a real-time human behavior prediction model that uses multi-resolution features.
    In the high-resolution input, the model observes the Manhattan distance difference between the agent and each potential goal during recent behavior, which is robust to obtain the agent's short-term intention.
    The low-resolution historical state matrix effectively organizes the long-term memory and helps the model extract the high-level feature.
    In addition, the supervised learning-based training provides a straightforward and automatic way to organize and learn the internal correlations from the human subjects data.
    After training, the experimental results demonstrated that our method is robust and accurate at effectively utilizes prior knowledge to predict human behavior.

%
%
%
%
    
\end{document}